\documentclass[
]{ceurart}

\usepackage{listings}
\usepackage{subfig} 

\usepackage{amsthm}
\usepackage{amssymb} 
\usepackage{mathtools} 

\newtheorem{definition}{Definition}[section]

\newtheorem{example}{Example}[section]

\DeclareMathOperator{\agg}{agg}

\DeclarePairedDelimiterX{\infdivx}[2]{(}{)}{%
  #1\;\delimsize\|\;#2%
}

\usepackage[size=footnotesize]{todonotes}

\begin{document}

\copyrightyear{2024}
\copyrightclause{Copyright for this paper by its authors.
  Use permitted under Creative Commons License Attribution 4.0
  International (CC BY 4.0).}

\conference{AEQUITAS 2024: Workshop on Fairness and Bias in AI | co-located with ECAI 2024, Santiago de Compostela, Spain}

\title{Measuring and Mitigating Bias for Tabular Datasets with Multiple Protected Attributes}

\tnotemark[1]

\author[1]{Manh Khoi Duong}[%
orcid=0000-0002-4653-7685,
email=manh.khoi.duong@hhu.de,
]
\cormark[1]
\address[1]{Heinrich Heine University, Universit\"atsstra\ss{}e 1, 40225 D\"usseldorf, Germany}

\author[1]{Stefan Conrad}[%
orcid=0000-0003-2788-3854,
email=stefan.conrad@hhu.de,
]

\cortext[1]{Corresponding author.}

\begin{abstract}
Motivated by the recital (67) of the current corrigendum of the
AI Act in the European Union, we propose and present
measures and mitigation strategies for discrimination in tabular datasets.
We specifically focus on datasets that contain multiple protected attributes, such
as nationality, age, and sex.
This makes measuring and mitigating bias more challenging, as
many existing methods are designed for a single protected attribute.
This paper comes with a twofold contribution:
Firstly, new discrimination measures are introduced.
These measures are categorized in our framework along with existing
ones, guiding researchers and practitioners in choosing the
right measure to assess the fairness of the underlying dataset.
Secondly, a novel application of an existing bias mitigation
method, \texttt{FairDo}, is presented.
We show that this strategy can mitigate any type of discrimination,
including intersectional discrimination,
by transforming the dataset.
By conducting experiments on real-world datasets (Adult, Bank, COMPAS),
we demonstrate that de-biasing datasets with multiple protected attributes is
possible.
All transformed datasets show a reduction in discrimination,
on average by 28\%.
Further, these datasets do not compromise
any of the tested machine learning models' performances significantly
compared to the original datasets.
Conclusively, this study demonstrates the effectiveness of the
mitigation strategy used and contributes to the ongoing discussion
on the implementation of the European Union's AI Act.
\end{abstract}

\begin{keywords}
  Machine Learning \sep
  Bias Mitigation \sep
  Intersectional Discrimination \sep
  Fairness \sep
  AI Act
\end{keywords}

\maketitle

\section{Introduction}
\begin{figure}
    \centering
    \includegraphics[width=0.6\textwidth]{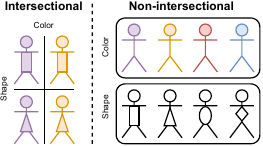}
    \caption{Stick figures can be differentiated
    by their color and shape. In intersectional discrimination, attributes are intersected, which leads to new subgroups. In non-intersectional, each attribute is treated independently, i.e., colors and shapes are not intersecting in this case.}
    \label{fig:overview}
\end{figure}

Discrimination in artificial intelligence (AI) applications
is a growing concern since the adoption of the \emph{AI Act}
by the European Parliament on March 13, 2024~\cite{eu2024aiactcorrigendum}.
It still remains a significant challenge across
numerous domains~\citep{liobait2017MeasuringDI,zafar2017-disparate, corbett2017conditionalstat,barocas-hardt-narayanan2019book}.
To prevent biased outcomes, \emph{pre-processing} methods
are often used to mitigate biases in datasets before training machine learning models~\citep{feldman2015certifying,fairlearn,agarwal2018reductions,duong2024framework}.
The current corrigendum of the \emph{AI Act}~\cite{eu2024aiactcorrigendum} emphasizes this in Recital (67):\begin{quote}
    \emph{``[...] The data sets should also have the appropriate statistical properties, including as regards the persons or groups of persons in relation to whom the high-risk AI system is intended to be used, with specific attention to the mitigation of possible biases in the data sets [...]''}
\end{quote}
Since datasets often consist of multiple protected attributes,
pre-processing methods should be able to handle these cases.
However, only a few works have addressed this
issue~\citep{fairlearn,fouldsbayesian2020,yang2020overlapping,noisypacelis21a,kang2021infofair}
and de-biasing such datasets
is still an ongoing research topic.
In addition, there is no straightforward approach to managing multiple protected attributes, as shown in Figure~\ref{fig:overview}.

Our paper mainly focuses on how to
measure and mitigate discrimination in datasets where
multiple protected attributes are present.
In our first contribution, we provide a comprehensive
categorization of discrimination measuring methods.
Besides introducing new measures for some of these cases,
we also categorize existing measures from the literature.
Some of the listed measures specifically address
\emph{intersectional discrimination} and \emph{non-binary groups}.
The second contribution deals with bias mitigation.
For this, we use our published
pre-processing framework, \texttt{FairDo}~\cite{duong2024framework},
that is \emph{fairness-agnostic}.
The fairness-agnostic property makes it possible to
define any discrimination measure that should be minimized.
By implementing the introduced measures, we can
therefore mitigate biases for multiple protected attributes.
Another advantage of \texttt{FairDo} is that it
preserves data integrity and does not modify
the features of individuals during the optimization process,
unlike other methods~\cite{zemel2013learning,zafar2017-disparate,fairlearn}.

We evaluated our methodology on popular tabular datasets
with fairness concerns, such as Adult~\cite{ron1996_adult},
Bank~\cite{moro2014bank}, and COMPAS~\cite{larson_angwin_mattu_kirchner_2016}.
We used different discrimination measures to evaluate the
effectiveness of the bias mitigation process.
Because a successful mitigation process does not
guarantee that the outcomes of machine learning models are fair,
we trained machine learning models on the transformed datasets and
evaluated their predictions regarding fairness and performance.
The code for the experiments can be found in the
accompanying repository: \url{https://github.com/mkduong-ai/fairdo/evaluation}.

The results of the
bias mitigation process as well as the
performance of the machine learning models are promising.
They indicate that achieving fairness in datasets
with multiple protected attributes is possible, and
\texttt{FairDo} is a proper framework for this task.
Overall, our work contributes technical solutions
for stakeholders to enhance the fairness of datasets and machine learning models,
aiming for compliance with the \emph{AI Act}~\cite{eu2024aiactcorrigendum}.

\section{Preliminaries}\label{section:notation}
To handle multiple protected attributes,
we define $\mathcal{Z} = \{Z_1, \ldots, Z_p \}$ as a set of protected attributes.
It can represent the set of sociodemographic features such as age, gender, and ethnicity.
These factors may make individuals vulnerable to discrimination.
Each protected attribute $Z_k \in \mathcal{Z}$ is formally a
\emph{discrete random variable} that can take on values from
the sample space $g_k$.
In this context, we refer $g_k$ to groups that describe
distinct social categories of a protected attribute.
For example, let $Z_k$ represent gender; then $g_k$ is a set containing the genders male, female, and non-binary.
To avoid limitations to a particular group fairness notion, we introduce a generalized notation based
on the works of \citet{liobait2017MeasuringDI},
\citet{duong2024framework} in the following.

\begin{definition}[Treatment]\label{def:treatment}
	Let $E_1, E_2$ be events and $Z_k$ be a random variable that can
	take on values from $g_k$, then we call the conditional probability
	\begin{equation*}
		P(E_1 \mid E_2, Z_k=i)
	\end{equation*}
	\emph{treatment}, where $i \in g_k$.
	$E_1$ describes some favorable outcome, 
	such as getting accepted for a job,
	while $E_2$ often represents some additional information
	about the individual, such as their qualifications.
\end{definition}

\begin{definition}[Fairness Criteria]\label{def:fairness}
	With the definition of treatment, we can define
	fairness criteria that demand equal treatment for different groups.
	Let $P(E_1 \mid E_2, Z_k=i)$ and $P(E_1 \mid E_2, Z_k=j)$ be treatments, then we call the following equation:
	\begin{equation*}
		P(E_1 \mid E_2, Z_k=i) = P(E_1 \mid E_2, Z_k=j)
	\end{equation*}
	a \emph{fairness criterion}, for all $i, j \in g_k$.
\end{definition}
\noindent Definition~\ref{def:fairness} allows us to define various group fairness criteria, including \emph{statistical parity}~\citep{calders2009}, \emph{predictive parity}~\citep{zafar2017-disparate}, \emph{equality of opportunity}~\citep{hardt2016equality}, etc.
They all demand some sort of equal outcome
for different groups and can be defined
by configuring the events $E_1, E_2$.
For instance, statistical parity~\cite{calders2009}
requires that two different groups have an equal probability of receiving a favorable outcome $(Y=1)$.
\begin{example}[Statistical Parity~\citep{calders2009}]
	To define statistical parity for the attribute $Z_k$ using our notation, we set $E_1 \coloneqq (Y=1)$ and $E_2 \coloneqq \Omega$.
	By setting $E_2$ to the sample space $\Omega$,
	we compare the probabilities of the event $Y=1$ across different groups without conditioning on any additional event:
    \begin{align*}
		P(Y=1 \mid \Omega, Z_k=i) &= P(Y=1 \mid \Omega, Z_k=j)\\
        \iff P(Y=1 \mid Z_k=i) &= P(Y=1 \mid Z_k=j),
    \end{align*}
    where $i, j \in g$ represent different groups.
\end{example}
\noindent In real-world applications, achieving equal probabilities for
certain outcomes is not always possible.
Due to variations in sample sizes in the groups,
it is common to yield unequal treatments,
even when they are similar.
Thus, existing literature~\cite{liobait2017MeasuringDI}
uses the absolute difference to quantify the strength of
discrimination.

\begin{definition}[Disparity]\label{def:disparity}
	Let $P(E_1 \mid E_2, Z_k=i)$ and $P(E_1 \mid E_2, Z_k=j)$ be two treatments, then we refer to
	\begin{equation*}
		\delta_{Z_k}(i, j, E_1, E_2) = |P(E_1 \mid E_2, Z_k=i) - P(E_1 \mid E_2, Z_k=j)|
	\end{equation*}
	as the \emph{disparity}, for all $i, j \in g_k$.
	Trivially, $\delta_{Z_k}$ is commutative regarding $i, j$.
	In practice, it prevents reverse discrimination due to the absolute value.
\end{definition}

\begin{definition}[Discrimination]\label{def:disc}
	We use $\psi \colon \mathbb{D} \rightarrow \mathbb{R}$ to denote some discrimination measure
	that quantifies the discrimination inherent in any dataset
	$\mathcal{D} \in \mathbb{D}$.
	A dataset $\mathcal{D}$ consists of features,
	protected attributes, and labels for each individual.
	The explicit form of $\psi$ depends on the cases introduced
	in Section~\ref{section:measuredisc}.
\end{definition}

\section{Measuring Discrimination for Multiple Attributes}\label{section:measuredisc}
We found that numerous scenarios arise when dealing with multiple protected attributes.
We categorize these scenarios based on the number of groups, denoted as $|g|$,
and the number of protected attributes, denoted as $|\mathcal{Z}|$.
By going through all cases, we present possible approaches from the literature
as well as our own suggestions to measure discrimination.

\subsection{Single Protected Attribute $(|\mathcal{Z}| = 1)$}
In the case of having only one protected attribute, i.e., $|\mathcal{Z}| = |\{Z_1\}| = 1$,
we distinguish between cases by the number of available groups
$|g|$ in the dataset.
We categorize the cases by $|g| = 0, 1, 2$, and $|g| > 2$.
\subsubsection{No Groups ($|g| = 0$)}
When there are no groups, the measurement of discrimination is impossible if no assumptions are being made.
Discrimination can be assessed through proxy variables~\cite{mehrabi2021survey}; however, this approach can be imprecise
and may introduce new biases.
This case is equivalent to having no protected attribute, i.e.,
$|\mathcal{Z}| = 0$.

\subsubsection{Single Group ($|g| = 1$)}
Similarly to the case of having no groups,
discrimination cannot be measured when having only one group.
For this, we propose practices where prior information can be incorporated:
\begin{enumerate}
	\item \emph{No discrimination}:
	As no difference towards any other group can be measured,
	returning a discrimination score of 0 is one viable option.
	\begin{equation}
		\psi(\mathcal{D}) = 0.
	\end{equation}
	\item\label{case:diffopttreat} \emph{Difference to optimal treatment}:
	Another way is to return the absolute difference
	of the group's outcome to the optimal treatment. For example, group $i$
	has an 80\% chance of receiving the favorable treatment.
	Ideally, having a 100\% chance
	would represent the optimal scenario.
	Therefore, the discrimination score is 20\% in this case.
	It is given by:
	\begin{equation}
		\psi(\mathcal{D}) = |P(E_1 \mid E_2, Z_1=i) - 1|.
	\end{equation}
	\item \emph{Difference to expected treatment}:
	We can use the expected treatment as a reference point.
	For example, we know that a company has a 50\% acceptance rate
	for job applications.
	Now a machine learning classifier is trained to predict
	whether an applicant will be accepted
	and the model's predictions result in a 60\% acceptance rate for group $i$.
	Hence, the model is positively biased towards group $i$
	by 10\%. This can be formulated as:
	\begin{equation}
		\psi(\mathcal{D}) = |P(E_1 \mid E_2, Z_1=i) - p_\text{expect.}|,
	\end{equation}
	where $p_\text{expect.}$ is the expected treatment.
	It can describe the average treatment across all groups~\cite{pmlr-v80-kearns18a} or some other prior information that is
	not included in the dataset.
\end{enumerate}

\subsubsection{Binary Groups ($|g| = 2$)}
Without using any prior information,
we can calculate the discrimination score by taking the
absolute difference between the treatments of the two groups,
as advised by Žliobaitė~\cite{liobait2017MeasuringDI}.
The discrimination measure $\psi$ is then simply given by the
disparity as mentioned in
Definition~\ref{def:disparity}.

\subsubsection{Non-binary Groups ($|g| > 2$)}
While the case for binary attributes is straightforward,
it becomes non-trivial for non-binary attributes that arise
naturally in real-world data.
We can fall back to $|g| = 2$ by calculating the absolute difference between every distinct group $i, j \in g$.
Because the discrimination between $i$ and $j$ is the same as between $j$ and $i$, only $\binom{|g|}{2}$ pairs need to be compared
and we use an aggregation function $\agg^{(1)}$ to report the differences~\cite{liobait2017MeasuringDI}.
\citet{lum2022debiasing} refers to measures that aggregate
or summarize
discrimination scores as \emph{meta-metrics}.
The aggregate can be the \emph{sum} or \emph{maximum} function, depending on the use case.
The result for a single protected attribute $Z_k$ with two or more groups can be computed as follows:
\begin{equation}\label{eq:aggnonbinary}
	\psi(\mathcal{D}) = \underset{i, j \in g_k, i < j}{\agg^{(1)}} \delta_{Z_k}(i, j, E_1, E_2),
\end{equation}
where $\delta_{Z_k}$ is the disparity as defined in Definition~\ref{def:disparity} and $i < j$ ensures that each pair is considered only once (assuming label-encoded groups).
According to Žliobaitė~\cite{liobait2017MeasuringDI}
and her personal discussions with legal experts, she
advocates using the maximum function, i.e.,
\begin{align}
	\psi(\mathcal{D}) &= \underset{i, j \in g_k, i < j}{\max} \delta_{Z_k}(i, j, E_1, E_2) \label{eq:aggnonbinarymax} \\
	&=  \underset{i \in g_k}{\max} \, P(E_1 \mid E_2, Z_k=i) - \underset{j \in g_k}{\min} \, P(E_1 \mid E_2, Z_k=j). \label{eq:aggnonbinarymax2}
\end{align}
Equation~\eqref{eq:aggnonbinarymax} describes
the maximum discrimination obtainable between two groups.
An alternative and equivalent formulation is given in Equation~\eqref{eq:aggnonbinarymax2}~\cite{fairlearn}.
The latter is computationally more efficient as it requires $\mathcal{O}(2|g|)$ operations compared to $\mathcal{O}(|g|^2)$ operations for the former.

A more general approach to measuring discrimination
is to calculate some form of \emph{correlation coefficient} between the protected attribute and the outcome.
The correlation coefficient can be calculated using
Pearson's correlation~\cite{pearsonr}, Spearman or Kendall's rank correlation~\cite{spearman1904,kendall1938}.
The discrimination measure can then be defined as the absolute value of the correlation coefficient:
\begin{equation}
	\psi(\mathcal{D}) = |\text{Corr}(E_1, Z_k)|.
\end{equation}
This approach can be applied to
any number of groups.
\texttt{Fairlearn} provides a pre-processing method
that removes the correlation between the protected attribute
and the outcome by transforming the data~\cite{fairlearn}.
However, the given approach violates data integrity constraints
as categorical attributes are transformed into continuous values.
Moreover, zero correlation does not imply independence
between two variables.

\subsection{Multiple Protected Attributes $(|\mathcal{Z}| > 1)$}
There are several ways to measure discrimination for
multiple protected attributes
($|\mathcal{Z}| > 1$). Based on the works of~\citet{pmlr-v80-kearns18a},
\citet{yang2020overlapping} and \citet{kang2021infofair},
we categorize them into two approaches: \emph{intersectional} and \emph{non-intersectional} (see Figure~\ref{fig:overview}). Intersectional approaches consider the intersection of identities.
The overlapping of such identities forms \emph{subgroups}~\cite{pmlr-v80-kearns18a}.
Non-intersectional approaches
treat each protected attribute independently~\cite{yang2020overlapping}.

\subsubsection{Intersectional Discrimination}
The central idea of intersectionality is that individuals experience
overlapping forms of oppression or privilege
based on the combination of multiple social categories they belong to.
In the following, we will introduce definitions to
formulate intersectional discrimination,
which is based on the work of \citet{pmlr-v80-kearns18a}.

\begin{definition}[Subgroup~\cite{pmlr-v80-kearns18a}]\label{def:subgroup}
	Let $\mathcal{Z} = \{Z_1, \ldots, Z_p\}$ be a set of discrete random variables
	representing protected attributes that can take on values from
	corresponding groups $g_1, \ldots, g_p$.
	A \emph{subgroup} $i$ is defined as $i = (i_1, \ldots, i_p) \in g_1 \times \ldots \times g_p$.
	In other words, a \emph{subgroup}
	encompasses multiple groups from different protected attributes.
\end{definition}

\begin{definition}[Subgroup Treatment]\label{def:subgrouptreatment}
	Let $i$ be a subgroup as defined in Definition~\ref{def:subgroup}
	and let $\mathcal{Z} = \{Z_1, \ldots, Z_p\}$ be a set of discrete random variables.
	\emph{Subgroup treatment} is then defined as:
	\begin{equation*}\label{eq:conditionaldef}
		P(E_1 \mid E_2, Z_1=i_1, \ldots, Z_p=i_p).
	\end{equation*}
\end{definition}

\begin{definition}[Subgroup Disparity]\label{def:subgroupdisparity}
	Let $\mathcal{Z} = \{Z_1, \ldots, Z_p\}$ be a set of discrete random variables.
	Let $i, j \in g_1 \times \ldots \times g_p$ be two subgroups
	with $i = (i_1, \ldots, i_p)$ and $j = (j_1, \ldots, j_p)$.
	The disparity between two subgroups is denoted as $\hat{\delta}_{\mathcal{Z}}$
	and is given by:
	\begin{equation*}
		\hat{\delta}_{\mathcal{Z}}(i, j, E_1, E_2) = |P(E_1 \mid E_2, Z_1=i_1, \ldots, Z_p=i_p) - P(E_1 \mid E_2, Z_1=j_1, \ldots, Z_p=j_p)|.
	\end{equation*}
\end{definition}
\noindent Similarly to Equation~\eqref{eq:aggnonbinary},
we can calculate the discrimination score for multiple protected attributes
by aggregating disparities across all subgroups.
A subgroup can be treated like a normal group.
According to Definition~\ref{def:subgroup},
there are theoretically at least
$2^p$ subgroups, where $p$ is the number of protected attributes.
However, not all subgroups may be available in the dataset.
For unavailable subgroups, the disparity cannot be calculated
as the corresponding treatment is undefined.

Let us denote the set of available subgroups as $G_\text{avail} \subseteq g_1 \times \ldots \times g_k$.
To finally capture the discrepancies across all available subgroup pairs,
an aggregation function $\agg^{(1)}$ is applied to the subgroup disparities $\hat{\delta}_{\mathcal{Z}}$:
\begin{align}\label{eq:aggmultiintersecting}
	\psi_\text{intersect}(\mathcal{D}) &= \underset{i, j \in G_\text{avail}}{\agg^{(1)}} \hat{\delta}_{\mathcal{Z}}(i, j, E_1, E_2).
\end{align}
Equation~\eqref{eq:aggmultiintersecting} represents the aggregated
discrimination between all available subgroups in the dataset.
When using the maximum function as the aggregator, the
calculations are equivalent to Equation~\eqref{eq:aggnonbinarymax}
and Equation~\eqref{eq:aggnonbinarymax2}.
The only difference is that the conditionals are now subgroups
instead of groups:
\begin{align}
	\psi_\text{intersect}(\mathcal{D}) &= \underset{i, j \in G_\text{avail}}{\max} \hat{\delta}_{Z_k}(i, j, E_1, E_2) \label{eq:aggintersectmax} \\
	&=  \underset{i \in G_\text{avail}}{\max} \, P(E_1 \mid E_2, Z_1=i_1, \ldots, Z_p=i_p) - \underset{j \in G_\text{avail}}{\min} \, P(E_1 \mid E_2, Z_1=j_1, \ldots, Z_p=j_p). \nonumber \label{eq:aggintersectmax2}
\end{align}
\noindent \citet{kang2021infofair} also dealt with intersectional
discrimination in their work by
introducing a multivariate random variable $Z$
where each dimension represents a protected attribute.
Their fairness objective is to minimize the mutual information
between the outcome and the multivariate random variable.
By minimizing the mutual information, the outcome is
independent of the protected attributes, which is a
desirable property for fairness~\cite{zemel2013learning,ghassami2018information}.
In this context, zero mutual information implies 
the absence of intersectional discrimination~\cite{kang2021infofair}.
However, this approach relies on expensive
techniques to approximate the mutual information.
Using our notation, their formulation can be written as~\cite{kang2021infofair}:
\begin{equation}
	\psi_\text{MI}(\mathcal{D}) = \text{MI}(E_1, Z),
\end{equation}
where $\text{MI}$ denotes the mutual information.

\begin{table}[t]
	\centering
	\caption{Example dataset of individuals receiving a favorable ($Y=1$) or unfavorable ($Y=0$) outcome.
		The dataset shows four individuals with their respective
		age group and sex.}
	\label{tab:example}
	\begin{tabular}{lllr}
		\toprule
		\textbf{Individual} & \textbf{Age} & \textbf{Sex} & \textbf{Outcome ($Y$)}\\
		\midrule
		1 & Old & Male & 1 \\
		2 & Old & Female & 0 \\
		3 & Young & Male & 0 \\
		4 & Young & Female & 1 \\
		\bottomrule
	\end{tabular}
\end{table}

\subsubsection{Non-intersectional Discrimination}
The problem with measuring discrimination for intersectional groups
is that it has an upward bias
when using meta-metrics~\cite{lum2022debiasing}.
This is because the number of subgroups grows exponentially
with the number of protected attributes.
This leads to many subgroups where the number of samples
in each subgroup is possibly small, resulting in larger noise
in the treatment estimates~\cite{lum2022debiasing}.

Besides intersectional groups, \citet{yang2020overlapping}
listed a non-intersectional definition of groups,
called \emph{independent groups}.
Building on the definition of \emph{independent groups},
we propose an appropriate approach to measure discrimination
for this type of groups.
It is more suitable when dealing with a large number of subgroups
or when intersectional discrimination is not deemed important.
Our non-intersectional approach treats each protected attribute
independently and aggregates the discrimination scores across
all protected attributes.
For this, a second aggregate function
with $\agg^{(2)}$ is introduced, yielding the following equation:
\begin{equation}\label{eq:aggmulti}
	\psi_\text{indep}(\mathcal{D}) = \underset{Z_k \in \mathcal{Z}}{\agg^{(2)}} \left\{ \underset{i, j \in g_k, i < j}{\agg^{(1)}} \delta_{Z_k}(i, j, E_1, E_2) \right\}.
\end{equation}
The first-level aggregator $\agg^{(1)}$ aggregates disparities
within a protected attribute, considering unique pairs of groups
$i$ and $j$.
The second-level aggregator $\agg^{(2)}$ then combines the results
across all protected attributes.
By applying both operators, we obtain a discrimination measure that
captures disparities between groups across multiple attributes.

\subsubsection{Example}
Let us consider a dataset with two protected attributes, age and
sex (see Table~\ref{tab:example}).
The set of protected attributes is $\mathcal{Z} = \{Z_1, Z_2\} = \{\text{Age}, \text{Sex}\}$
and the set of available subgroups in the dataset is $G_\text{avail} = \{\text{Old, Young}\} \times \{\text{Male, Female}\}$.
We measure discrimination using \emph{statistical disparity}.
For simplicity, all aggregation functions are set to the maximum function.
The \emph{intersectional approach} yields the following discrimination score:
\begin{align}
	\psi_\text{intersect}(\mathcal{D}) &= \underset{i, j \in G_\text{avail}}{\max} \hat{\delta}_{\mathcal{Z}}(i, j, (Y=1), \Omega) \\
	&= \underset{i, j \in G_\text{avail}}{\max} \hat{\delta}_\text{\{Age, Sex\}}(i, j, (Y=1), \Omega) \nonumber \\
	&= \underset{i \in G_\text{avail}}{\max} P(Y=1 \mid Z_1=i_1, Z_2=i_2) - \underset{j \in G_\text{avail}}{\min} P(Y=1 \mid Z_1=j_1, Z_2=j_2) \nonumber \\
	&= |P(Y=1 \mid \text{Age}=\text{Old}, \text{Sex}=\text{Male}) - P(Y=1 \mid \text{Age}=\text{Young}, \text{Sex}=\text{Male})| = 1, \nonumber
\end{align}
while the discrimination score for the \emph{non-intersectional approach} is given by:
\begin{align}
	\psi_\text{indep}(\mathcal{D}) &= \underset{Z_k \in Z}{\max} \left\{ \underset{i, j \in g_k, i < j}{\max} \delta_{Z_k}(i, j, (Y=1), \Omega) \right\} \\
	&= \max \left\{\delta_{\text{Age}}(\text{Old}, \text{Young}, (Y=1), \Omega),
	\delta_{\text{Sex}}(\text{Male}, \text{Female}, (Y=1), \Omega)\right\} \nonumber \\
	&= \max \{|0.5 - 0.5|, |0.5 - 0.5|\} = \max \{0, 0\} = 0. \nonumber
\end{align}
The non-intersectional approach yields a discrimination score of 0 because
the disparities for both protected attributes are 0.
This is quite different from the intersectional approach, which
reports a discrimination score of 1. As seen,
the results can differ depending on the approach.
\section{Experiments}
Our experimentation follows a pipeline consisting of \emph{data pre-processing}, \emph{bias mitigation}, \emph{model training}, and \emph{evaluation}.
To mitigate bias in tabular datasets with multiple protected attributes, we used the sampling method, \texttt{FairDo}~\cite{duong2024framework}, that constructs fair datasets by selectively sampling data points.
The method is very flexible and only requires the user to
define the discrimination measure that should be minimized.
In our case, we are interested in a dataset that has minimal
bias across multiple protected attributes.
The experiments revolve around the following research questions:
\begin{itemize}
    \item \textbf{RQ1} Is it possible to yield a fair dataset with \texttt{FairDo}, where bias for multiple protected attributes is reduced?
    \item \textbf{RQ2} Are machine learning models trained on fair datasets more fair in their predictions
    than those trained on original datasets?
\end{itemize}

\subsection{Experimental Setup}
\paragraph{Datasets and Pre-processing}
\begin{table}[tb]
    \caption{Overview of Datasets}
    \centering
    \label{table:dataset_comparison}
    \begin{tabular}{lrrp{1.5cm}p{4.7cm}p{3cm}}
    \hline
    \textbf{Dataset} & \textbf{Samples} & \textbf{Feats.} & \textbf{Label} & \textbf{Protected Attributes} & \textbf{Description}\\
    \hline
    Adult~\cite{ron1996_adult} & 32\,561 & 21 & Income & \textbf{Race}: White, Black, Asian-Pacific-Islander, American-Indian-Eskimo, Other\newline \textbf{Sex}: Male, Female & Indicates individuals earning over \$50,000 annually \\ \hline
    Bank~\cite{moro2014bank} & 41\,188 & 50 & Term deposit\newline subscription & \textbf{Job}: Admin, Blue-Collar, Technician, Services, Management, Retired, Entrepreneur, Self-Employed, Housemaid, Unemployed, Student, Unknown\newline
    \textbf{Marital Status}: Divorced, Married, Single, Unknown & Shows whether the client has subscribed to a term deposit.\\ \hline
    COMPAS~\cite{larson_angwin_mattu_kirchner_2016} & 7\,214 & 13 & 2-year\newline recidivism & \textbf{Race}: African-American, Caucasian, Hispanic, Other, Asian, Native American\newline
    \textbf{Sex}: Male, Female\newline
    \textbf{Age Category}: <25, 25-45, >45& Displays individuals that were rearrested for a new crime within 2 years after initial arrest.\\
    \hline
    \end{tabular}
\end{table}
The tabular datasets employed in our experiments include the Adult~\cite{ron1996_adult}, Bank~\cite{moro2014bank}, and COMPAS~\cite{larson_angwin_mattu_kirchner_2016} datasets.
They are known for their use in fairness research and contain multiple protected attributes.
We pre-processed the datasets by applying one-hot encoding
to categorical variables and label encoding to protected attributes.
Table~\ref{table:dataset_comparison} shows important characteristics of the datasets after pre-processing.

Each dataset was divided into training and testing sets
using an 80/20 split, respectively.
We ensured that the split was stratified (if possible)
based on protected attributes to maintain representativeness
across different groups in both sets.

\paragraph{Bias Mitigation}
Applying the bias mitigation method \texttt{FairDo}~\cite{duong2024framework} to the datasets 
can be regarded as a pre-processing step, too.
This is because the method simply returns a dataset
that is fair with respect to the given discrimination measure.
\texttt{FairDo}~\cite{duong2024framework} offers a variety of
options to mitigate bias, and we chose the \emph{undersampling}
method that removes samples.
In this option, the optimization
objective is stated as~\cite{duong2024framework}:
\begin{equation}
    \min_{\mathcal{D}_\text{fair} \subseteq \mathcal{D}} \quad \psi(\mathcal{D}_\text{fair}),
\end{equation}
where $\mathcal{D}$ is the training set of Adult, Bank, or COMPAS,
and $\psi$ is the fairness objective function.
We experimented with both $\psi_\text{intersect}$ and $\psi_\text{indep}$ as objectives functions.
Bias mitigation is only applied to the training set and the testing set remains unchanged.
\texttt{FairDo} internally uses genetic algorithms
to select a subset of the training set that minimizes the objective
function.
We used the same settings and operators as provided in the package and
only adjusted the population size (200) and the number of generations (400).

\paragraph{Model Training}
We utilized the \texttt{scikit-learn} library~\cite{scikit-learn} to train various machine learning classifiers, namely \emph{Logistic Regression} (LR), \emph{Support Vector Machine} (SVM), \emph{Random Forest} (RF), and \emph{Artificial Neural Network} (ANN). These classifiers were trained on both the original and fair datasets.
Classifiers trained on the original datasets serve
as a baseline for comparison.
We used the default hyperparameters given by
\texttt{scikit-learn} package for each classifier.

\paragraph{Evaluation Metrics}
We evaluated the models' predictions on fairness and performance
using the test set.
For fairness, we assessed $\psi_\text{intersect}$ and
$\psi_\text{indep}$.
For the classifiers' performances, we report
the \emph{area under the receiver operating characteristic curve} (AUROC)~\cite{Fawcett2006roc},
where higher values indicate better performances.
Because removing data points can compromise
the overall quality of the data,
we also report the number of subgroups before
and after bias mitigation to check for
representativeness.

\paragraph{Trials}
For each dataset and discrimination measure combination,
the bias mitigation process was repeated 10 times.
The results were averaged over the trials to obtain a more
robust evaluation.

\subsection{Results}
\paragraph{Fair Dataset Generation}
\begin{table}[tb]
    \caption{Average discrimination and number of subgroups
     before and after pre-processing the training sets with \texttt{FairDo}.}
     \label{table:results_training}
    \begin{tabular}{ll|rrrr}
    \hline
    Dataset & Metric & Disc. Before & \textbf{Disc. After} & Subgroups Before & Subgroups After \\
    \hline
    \multirow[t]{2}{*}{Adult} & $\psi_\text{indep}$ & $20\%$ & $13\%$ & 10 & 10 \\
     & $\psi_\text{intersect}$ & $31\%$ & $16\%$ & 10 & 10 \\
    \cline{1-6}
    \multirow[t]{2}{*}{Bank} & $\psi_\text{indep}$ & $24\%$ & $5\%$ & 48 & 48 \\
     & $\psi_\text{intersect}$ & $33\%$ & $15\%$ & 48 & 46.2 \\
    \cline{1-6}
    \multirow[t]{2}{*}{COMPAS} & $\psi_\text{indep}$ & $30\%$ & $5\%$ & 34 & 34 \\
     & $\psi_\text{intersect}$ & $100\%$ & $17\%$ & 34 & 28.8 \\
    \hline
    \end{tabular}
\end{table}

Table~\ref{table:results_training} shows the average discrimination before and after mitigating bias in the
training sets.
On all datasets, discrimination was reduced after applying \texttt{FairDo}.
Without considering group intersections, discrimination was reduced
by 7\%, 19\%, and 25\% for Adult, Bank, and COMPAS, respectively.
When considering intersectionality, the discrimination was reduced by 15\%, 18\%, and 83\%.
Hence, discrimination was reduced by 28\% on average across all datasets, thus answering \textbf{RQ1} positively.
When comparing the discrimination scores,
it can be observed that the intersectional discrimination scores
are generally higher.
This is because in the intersectional setting,
more subgroups are considered, which potentially leads to larger differences between them~\cite{pmlr-v80-kearns18a}.

We also report the number of subgroups before and after bias mitigation
to assess the impact of the undersampling method on the dataset.
The removal of subgroups can only be observed in the intersectional setting.
In the COMPAS dataset 5.2 out of 34 subgroups were removed on average,
indicating the largest amount of subgroups removed across all datasets.
While the Bank dataset consists of 48 subgroups, only 1.8 subgroups were removed on average.
Because the COMPAS dataset's initial intersectional discrimination score is 100\%,
removing more subgroups seems inevitable to reduce bias.

\paragraph{Model Performance and Fairness}
\begin{figure*}[tb]
    \centering
    \hspace{2.8em}
    \subfloat{\includegraphics[width=0.875\textwidth]{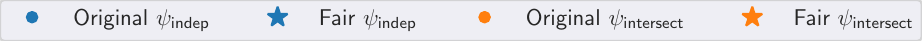}}\\
    \vspace{-0.45em}
    \addtocounter{subfigure}{-1}%
    \subfloat[Adult (LR)]{\includegraphics[height=0.111\textheight]{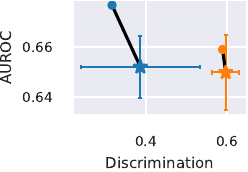}
    \label{fig:adult_lr}}%
    \subfloat[Adult (SVM)]{\includegraphics[height=0.111\textheight]{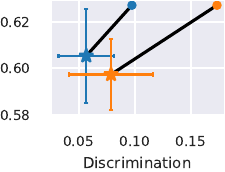}
    \label{fig:adult_svm}}%
    \subfloat[Adult (RF)]{\includegraphics[height=0.111\textheight]{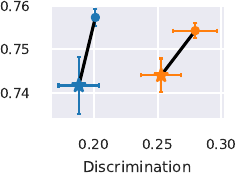}
    \label{fig:adult_rf}}
    \subfloat[Adult (ANN)]{\includegraphics[height=0.111\textheight]{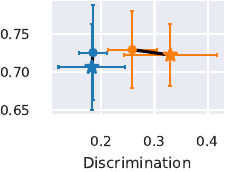}
    \label{fig:adult_ann}}

    \subfloat[Bank (LR)]{\includegraphics[height=0.111\textheight]{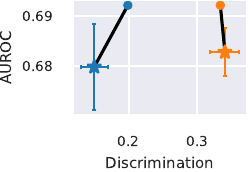}
    \label{fig:bank_lr}}%
    \subfloat[Bank (SVM)]{\includegraphics[height=0.111\textheight]{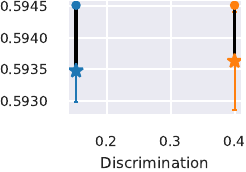}
    \label{fig:bank_svm}}%
    \subfloat[Bank (RF)]{\includegraphics[height=0.111\textheight]{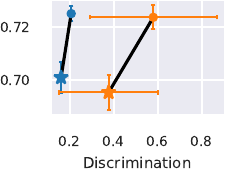}
    \label{fig:bank_rf}}
    \subfloat[Bank (ANN)]{\includegraphics[height=0.111\textheight]{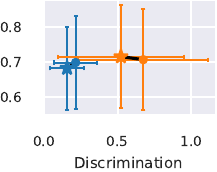}
    \label{fig:bank_ann}}
    
    \subfloat[COMPAS (LR)]{\includegraphics[height=0.111\textheight]{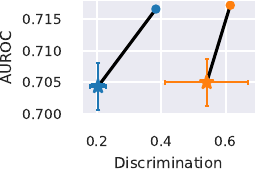}
    \label{fig:compas_lr}}%
    \subfloat[COMPAS (SVM)]{\includegraphics[height=0.111\textheight]{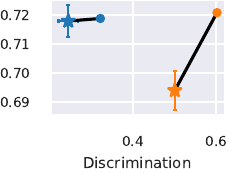}
    \label{fig:compas_svm}}%
    \subfloat[COMPAS (RF)]{\includegraphics[height=0.111\textheight]{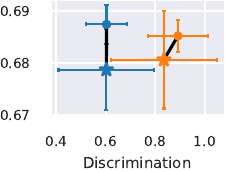}
    \label{fig:compas_rf}}
    \subfloat[COMPAS (ANN)]{\includegraphics[height=0.111\textheight]{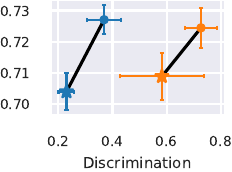}
    \label{fig:compas_ann}}
    \caption{Results on the test set. The x-axis represents the discrimination values (legend indicates used measure) and the y-axis represents the classifiers' performances.
    We compare the pre-processed (fair) data with the original data.
    The points/stars represent averages, and the error bars display
    the standard deviations
    of the AUROC and discrimination values over 10 trials.}
    \label{fig:results_test}
\end{figure*}

Figure~\ref{fig:results_test} shows the results of the classifiers'
performances on the test set.
The classifiers' performances are displayed on the y-axis, while
the discrimination values are shown on the x-axis.
We note that the axes do not share the same scale across the subfigures
for analytical purposes.

Classifiers trained on fair datasets
did not suffer a significant decline in performance
compared to those trained on original datasets.
In all cases, only a slight decrease of 1\%-3\% in performance can be noted.
This indicates that the bias mitigation process does not
compromise the dataset's fidelity and, therefore, the classifiers' performances.
Regarding discrimination, a significant reduction is evident.
The x-axis scales are much larger than the y-axis scales,
suggesting that changes in discrimination are larger than changes in performance.
For example, the RF classifier trained on the Bank dataset (Figure~\ref{fig:bank_rf})
shows a decrease in intersectional discrimination from 38\% to 15\%,
while the performance only decreases by 2\%.
Similar results can be observed for the other classifiers and datasets
as well, successfully addressing \textbf{RQ2}.
The results suggest that \texttt{FairDo} can be reliably used to
mitigate bias in tabular datasets for various measures that consider multiple protected attributes.
Still, we advise users to carefully perform similar analyses
when applying the method to their datasets.
\section{Discussion}
The results of our experiments show that the presented measures
detect discrimination in datasets with multiple protected attributes
differently.
When using the intersectional discrimination measure, more groups
are identified and compared to each other.
While subgroups are not ignored by this measure,
measuring higher discrimination scores by random chance becomes more likely~\cite{pmlr-v80-kearns18a,lum2022debiasing}.
In contrast, treating each protected attribute separately
prevents this issue but may lead to overlooking
discrimination.
The choice of measure
is up to the stakeholders and depends on the context of the dataset
and the regulations that apply to the AI system.
We generally recommend using the intersectional discrimination measure
if the number of individuals in each subgroup is large enough
to draw statistically significant conclusions.
Otherwise, treating each protected attribute separately is more
suitable.

By using the mitigation strategy
\texttt{FairDo}~\cite{duong2024framework}, the resulting
datasets in the experiments have improved statistical
properties regarding fairness.
Whether intersectionality was considered or not,
reducing discrimination in datasets was possible.
At the current state, the AI Act~\cite{eu2024aiactcorrigendum} does not
explicitly mention \emph{intersectional discrimination}
nor how to deal with multiple protected attributes generally.
While recital (67) states that datasets \emph{``should [...] have
the appropriate statistical properties''}, it does not specify
what these properties are.
Hence, our work serves as an initial guideline for
what these properties
could be and how to achieve them in practice.

\section{Conclusion}
Datasets often come with multiple protected attributes,
which makes measuring and mitigating discrimination more challenging.
Most existing studies only deal with a single protected attribute,
and works that consider multiple protected attributes
often focus on intersectionality.
In opposition to this, we proposed a new non-intersectional
measure that treats each protected attribute separately.
This is more suitable when the number
of subgroups is too large or the number of
individuals in each subgroup is small.
We used both intersectional and non-intersectional measures
as objectives and
applied the \texttt{FairDo} framework
to mitigate discrimination in multiple datasets.
The experiments show that discrimination was reduced
in all datasets and on average by 28\%.
Machine learning models trained on the bias-mitigated datasets
also improved their fairness while maintaining performance
compared to models trained on the original datasets.


\bibliography{references}




\end{document}